\documentclass[letterpaper, 10 pt, conference]{ieeeconf}  
\usepackage{cite}
\usepackage{picinpar}
\usepackage{amsmath}
\usepackage{url}
\usepackage{flushend}
\usepackage{colortbl}
\usepackage{soul}
\usepackage{multirow}
\usepackage{pifont}
\usepackage{color}
\usepackage{alltt}
\usepackage{hyperref}
\usepackage{siunitx}
\usepackage{breakurl}
\usepackage{epstopdf}
\usepackage{pbox}
\usepackage{subfigure}
\usepackage{threeparttable}
\usepackage{amssymb}
\usepackage{graphicx}
\usepackage{bm} 
\usepackage{adjustbox}
\usepackage{booktabs}
\usepackage[utf8]{inputenc}
\usepackage{booktabs}
\usepackage{threeparttable}
\usepackage{textcomp}
\usepackage{hyperref}
\usepackage{amsmath}

\usepackage{booktabs}
\usepackage{float}
\usepackage{indentfirst}
\usepackage{multirow}
\usepackage{multicol}
\usepackage[table,xcdraw]{xcolor}
\usepackage{colortbl}
\usepackage{makecell}
\usepackage{xcolor} 

\IEEEoverridecommandlockouts                              

\title{\LARGE \bf
LiDAR-VGGT: Cross-Modal Coarse-to-Fine Fusion for Globally Consistent and Metric-Scale Dense Mapping

}

\author{
        Lijie Wang,
        Lianjie Guo,
        Ziyi Xu,
        Qianhao Wang,
        Fei Gao,
	      Xieyuanli Chen
\thanks{L. Wang, L. Guo, Z. Xu, Q. Wang and F. Gao are with the State Key Laboratory of Industrial Control
Technology, Institute of Cyber-Systems and Control, Zhejiang University,
Hangzhou, 310027, China. (e-mail:
3210101760@zju.edu.cn)
}
\thanks{L. Wang, L. Guo, Q. Wang and F. Gao are also with Differential Robot Technology Co., Ltd., Hangzhou, China. This work was supported by Differential Robot Technology Co., Ltd.}
\thanks{X. Chen is with the College of Intelligence Science
and Technology, National University of Defense Technology,
Changsha 410007, China.}
\thanks{Corresponding author: Xieyuanli Chen(xieyuanli.chen@nudt.edu.cn).}
    } 

\begin{document}

\bstctlcite{IEEEexample:BSTcontrol}

\maketitle
\thispagestyle{empty}
\pagestyle{empty}

\begin{abstract}
Reconstructing large-scale colored point clouds is an important task in robotics, supporting perception, navigation, and scene understanding.
Despite advances in LiDAR inertial visual odometry (LIVO), its performance remains highly sensitive to extrinsic calibration. Meanwhile, 3D vision foundation models, such as VGGT, suffer from limited scalability in large environments and inherently lack metric scale.
To overcome these limitations, we propose LiDAR-VGGT, a novel framework that tightly couples LiDAR inertial odometry with the state-of-the-art VGGT model through a two-stage coarse-to-fine fusion pipeline: 
First, a pre-fusion module with robust initialization refinement efficiently estimates VGGT poses and point clouds with coarse metric scale within each session.
Then, a post-fusion module enhances cross-modal 3D similarity transformation, using bounding-box–based regularization to reduce scale distortions caused by inconsistent FOVs between LiDAR and camera sensors.
Extensive experiments across multiple datasets demonstrate that LiDAR-VGGT achieves dense, globally consistent colored point clouds and outperforms both VGGT-based methods and LIVO baselines.
The implementation of our proposed novel color point cloud evaluation toolkit will be released as open source.
\end{abstract}

\section{Introduction}
Reconstructing accurate colored point clouds serves as the foundation for key robotics capabilities, such as perception~\cite{slamfuture}, navigation~\cite{navigation}, and scene understanding~\cite{sceneunderstanding}, in complex environments. Moreover, large-scale 3D reconstruction is crucial for embodied navigation, enabling robots to safely explore unstructured environments~\cite{gaussnav}; for multi-robot collaboration, where shared maps support coordinated tasks~\cite{kimeramulti, lemonmapping}; and for autonomous driving, providing high-definition environmental models for localization and planning~\cite{driving}.

Classical LiDAR inertial visual odometry (LIVO) pipelines have achieved progress by fusing multi-modal measurements for trajectory estimation and map reconstruction~\cite{fastlivo, fastlivo2, r3live, r3live++, gslivo}. 
However, their performance heavily relies on precise extrinsic calibration and tight timestamp synchronization across sensors, which compromises robustness in real-world deployments.
Moreover, the inherent sparsity of LiDAR scans prevents the generation of dense, high-fidelity colored 3D reconstructions.

The rise of foundation models for 3D vision has opened new avenues for dense scene reconstruction. Early works such as DUSt3R~\cite{dust3r} and MASt3R~\cite{mast3r} demonstrated the potential to directly predict dense point clouds from images, though they are limited to processing image pairs and rely on post-fusion for larger scenes. More recently, the Visual Geometry Grounded Transformer (VGGT)~\cite{vggt} extended this paradigm by jointly performing pose estimation, depth prediction, and dense 3D reconstruction from multiple uncalibrated RGB frames in a single inference. Although VGGT achieves good local reconstructions without explicit optimization, its scalability is restricted by GPU memory, and the generated colored point clouds lack metric scale, limiting its practicality for large-scale robotic applications. 
To process long image sequences in large environments, recent works~\cite{vggt-long, vggt-slam} 
adopt segments processing by dividing images into smaller sessions and aligning adjacent sessions to achieve large-scale mapping. Although this reduces memory requirements, such pairwise alignment only enforces relative scale between submaps without establishing a globally consistent metric scale. 
Furthermore, these methods rely heavily on point cloud correspondences derived from pixel correspondences in overlapping regions, making them fragile in low-overlap scenarios.

\begin{figure}[!t]
  \centering
  \includegraphics[width=1.0\linewidth]{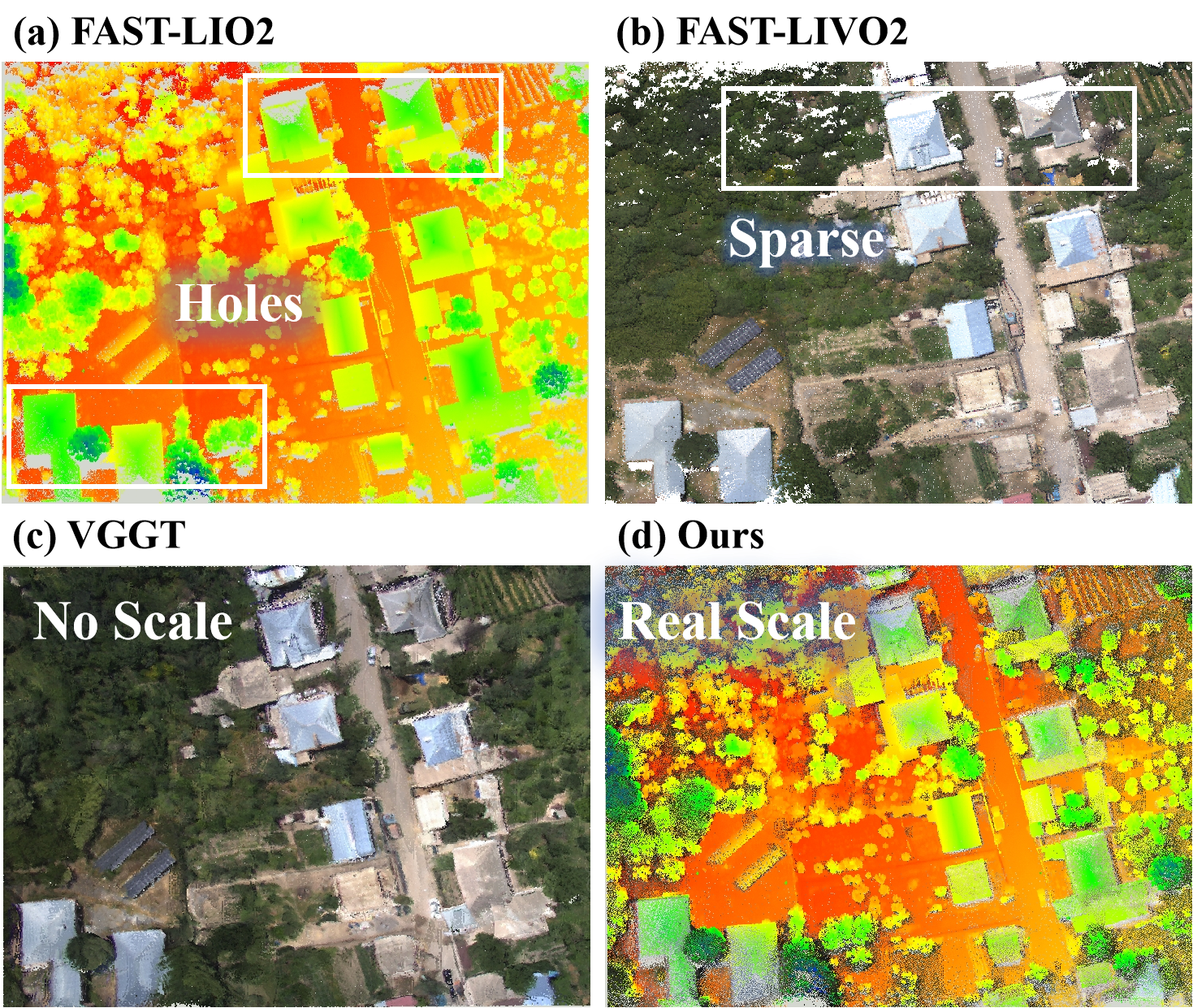}
   \vspace{-20pt}
  \caption{Motivation figure: (a) and (b) present point clouds from FAST-LIO2~\cite{fastlio2} and FAST-LIVO2~\cite{fastlivo2}, which are sparse and contain holes due to limited LiDAR scanning coverage. (c) shows the dense yet scale-ambiguous map produced by VGGT~\cite{vggt}. (d) illustrates our method, which generates VGGT-based point clouds guided by LiDAR reconstruction, achieving dense maps with accurate metric scale.}
  \vspace{-10pt}
  \label{fig:registration}
\end{figure}

In general, feed-forward predictions such as VGGT generate dense and visually clean point clouds but lack metric scale. In contrast, LiDAR point clouds provide direct, high-precision depth measurements with real-world scaling and accurate geometry. When coupled with loop detection~\cite{ring++, overlapnet, fastloop} and pose graph optimization (PGO), they yield globally consistent structures and high-quality reconstructions. Therefore, by leveraging LiDAR to impose metric scale on VGGT’s reconstructions, we can obtain dense, clean, and metrically accurate colored maps.

To this end, we propose LiDAR-VGGT, a novel framework that tightly integrates LiDAR with VGGT through a coarse-to-fine fusion pipeline.
In the coarse fusion stage, LiDAR inertial odometry (LIO) poses are used to refine the VGGT poses via a robust refinement mechanism that combines scale RANSAC and linearity-based validation to reject outliers in the $Sim(3)$ transformation.
This enables the scale-normalized local poses of VGGT to be transformed into the real-world coordinate system while generating colored point clouds with approximate metric scale.
Then the fine fusion module leverages the initial value to further improve the accuracy and consistency of the colored map. Each session’s VGGT point cloud is aligned with the LiDAR point cloud using an enhanced cross-modal $Sim(3)$ registration that employs bounding-box–based regularization to reduce scale distortions caused by field-of-view differences between the camera and LiDAR. This ensures stable registration and recovers the real-world metric scale. 
Finally, a global PGO is applied to eliminate divergence across different sessions and improve global consistency, enabling robust reconstruction of large-scale scenes.
For the evaluation of colored point clouds, existing methods mainly focus on rendering quality without considering ground truth maps~\cite{rised, gslivo}, and we propose a novel evaluation toolkit with four metrics to address this.

Our contributions can be summarized as follows.
\begin{itemize}
    \item We propose LiDAR-VGGT, to our knowledge the first system that fuses LiDAR with VGGT to achieve large-scale, metric-accurate, dense, and globally consistent colored point cloud reconstruction.
    \item We design a novel pre-fusion module that refines VGGT poses with LIO poses through linearity validation and scale-aware RANSAC, transforming them into the world coordinate system and producing colored point cloud with a coarse metric scale.
    \item We introduce a novel cross-modal $Sim(3)$ registration with bounding-box based regularization, which tackle the scale distortion caused by FOV inconsistency and robustly registers VGGT reconstructions of varying scales to LiDAR point cloud.
    \item We introduce a new tool for evaluating the color quality of large-scale, metric-accurate, dense point clouds. The full evaluation toolkit will be open-sourced\footnote{\url{https://github.com/NorwegianSmokedSalmon/Color-Map-Evaluation}}.

\end{itemize}

\section{Related Works}
\subsection{Multi-Modal SLAM and Reconstruction}

Classical Multi-Modal SLAM approaches fuse information from multiple sensors such as LiDAR, camera, and IMU to estimate trajectories and reconstruct 3D scenes. 
R$^3$LIVE \cite{r3live} and R$^3$LIVE++ \cite{r3live++} tightly couple the LiDAR point with the image pixel to achieve robust localization and RGB colorized point cloud reconstruction.
FAST-LIVO and FAST-LIVO2 \cite{fastlivo, fastlivo2} exploit the tight coupling between LiDAR points and image patches, enabling robust localization in degraded environments and producing high-precision 3D maps.
In parallel, recent advances in Neural Radiance Fields (NeRF) \cite{nerf} and 3D Gaussian Splatting (3DGS) \cite{3dgs} have inspired current works to integrate new map representations into SLAM systems.
Extensions such as LiDAR-NeRF \cite{lidarnerf} and SiLVR \cite{silvr} integrate LiDAR depth and SLAM trajectories to achieve scalable, geometrically accurate reconstructions with metric scale. Also, systems like Gaussian-LIC \cite{gaussianlic} and GS-LIVO \cite{gslivo} tightly fuse LiDAR, IMU, and visual data with Gaussian maps for real-time SLAM, maintaining robustness under challenging illumination, motion, and resource constraints.
In summary, these pipelines achieve accurate localization and high-precision reconstruction through different sensor fusion methods and map representation. However, they typically require accurate extrinsic calibration and precise time synchronization between sensors. Moreover, LiDAR provides only sparse range measurements, which restricts the density of reconstructions.

\subsection{Feed-forward Scene Reconstruction}
Beyond explicit geometric optimization, another line of research employs feed-forward neural networks to directly model 3D scenes from images. 
DUSt3R \cite{dust3r} first demonstrated the power of large-scale data-driven reconstruction, predicting dense point maps from image pairs and achieving scene reconstruction through global alignment. However, its geometric consistency degrades as the number of input images increases. Several follow-up works address scalability: Spann3R \cite{span3r} introduces a learned memory module for streaming image sequences, MV-DUSt3R+ \cite{tang2024mv} extends DUSt3R \cite{dust3r} to multi-frame inputs, and SLAM3R \cite{slam3r} leverages a Local-to-World module to incrementally align locally reconstructed point maps to a global coordinate system. While these methods enable incremental scene reconstruction, they lack loop-closure mechanisms to correct accumulated drift, and thus perform reliably only on short sequences. Fast3R \cite{Yang_2025_Fast3R} further scales feed-forward inference to larger image sets, but with limited accuracy.

Most recently, VGGT (Visual Geometry Grounded Transformer) \cite{vggt} represents a pioneering work that unifies multi-frame camera pose estimation, depth prediction, and dense reconstruction within a single feed-forward neural network. VGGT produces dense and visually clean reconstructions without explicit optimization, yet its scalability is fundamentally constrained by GPU memory, restricting inference to only tens of frames. To extend VGGT to large-scale mapping, VGGT-Long \cite{vggt-long} and VGGT-SLAM \cite{vggt-slam} partition long sequences into submaps and achieve large-scale mapping by pairwise registration with $Sim(3)$ and $SL(4)$ optimization. However, these segmented pipelines enforce only relative scale consistency. Due to the absence of a reliable global metric reference, frequently lead to curved planes and geometric distortions.
\section{Methodology}
\subsection{Problem Formulation and System Overview}
\begin{figure*}[!t]
  \centering
  \includegraphics[width=0.95\linewidth]{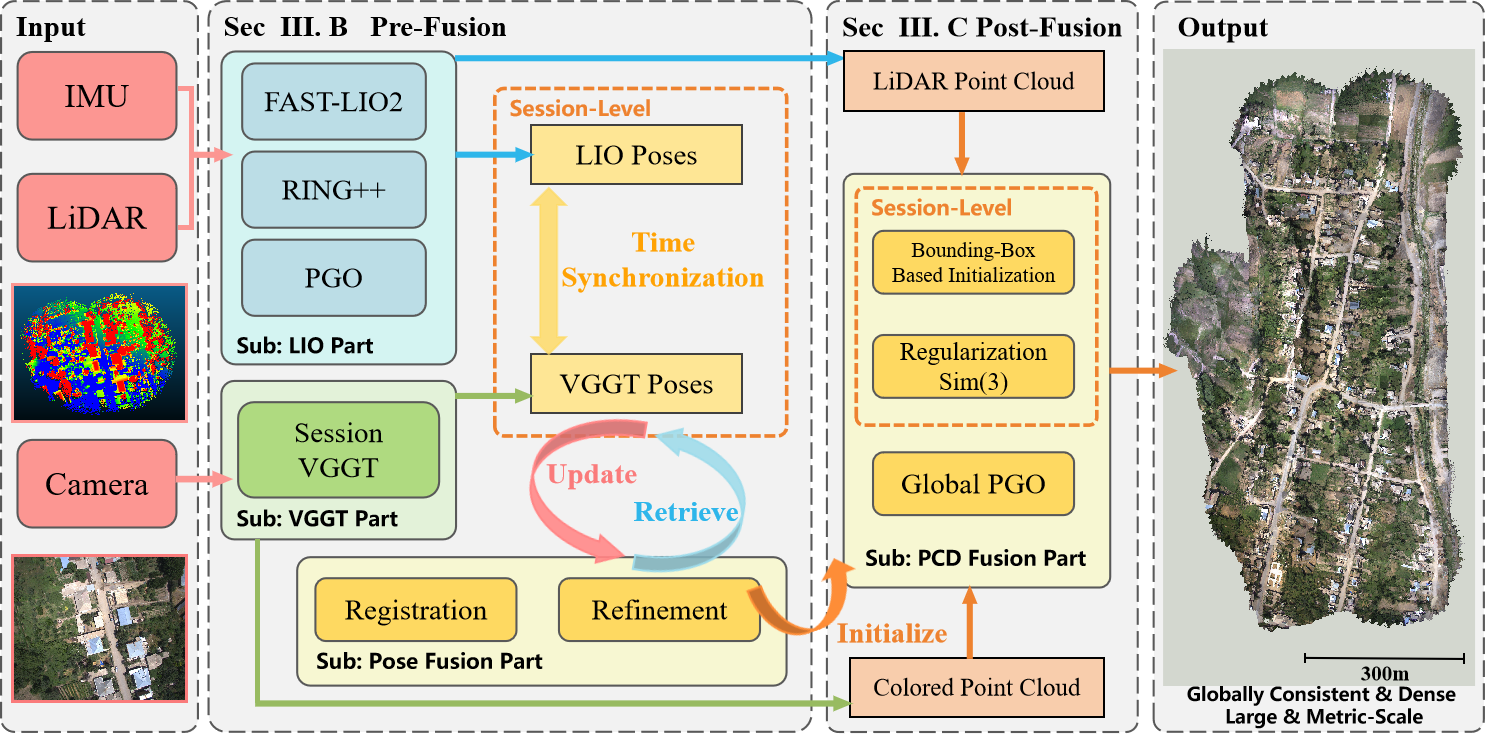}
   \vspace{-0.5cm}
  \caption{Overview of the proposed LiDAR-VGGT system. It takes multi-modal data from IMU, LiDAR, and camera as input and produces globally consistent, dense reconstructions with metric scale. Our pipeline follows a coarse-to-fine strategy, consisting of a pre-fusion module (Sec.~\ref{subsetcion:frontend}) and a post-fusion module (Sec.~\ref{subsection:backend}). The former fuses LIO and VGGT poses to establish an approximate real-world scale for VGGT-generated scene attributes, while the latter refines the VGGT-colored point cloud through enhanced cross-modal $Sim(3)$ registration and global pose graph optimization.} 
  \label{fig:system}
   \vspace{-0.2cm}
\end{figure*}

VGGT~\cite{vggt} takes a sequence of $N$ RGB images, $S_N = \{I_1, I_2,...,I_N \}$ as input, where each image is denoted as $I_i \in \mathbb{R}^{3 \times H \times W}$. It outputs the scene properties and reconstruction results, formulated as:
\begin{equation}
    f(S_N) = (\mathbf{g}_i, D_i, P_i, Tr_i)_{i = 1}^N.
\end{equation}

For image $I_i$, $\mathbf{g}_i \in \mathbb{R}^9$ represents the camera intrinsics and extrinsics, $D_i \in \mathbb{R}^{H \times W}$ is the predicted depth map, $P_i \in \mathbb{R}^{3 \times H \times W}$ is the corresponding point map, and $Tr_i \in \mathbb{R}^{C \times H \times W}$ is a grid of features for point tracking. 
VGGT and its extensions like VGGT-long and VGGT-SLAM still lack absolute metric scales in real-world scenarios, leaving large-scale metric reconstruction an open challenge.
In large-scale scenarios, the complete image sequence $S_N$ can be divided into $K$ overlapping sessions $ \mathbb{S}_K = \{ S_1, S_2, \dots, S_K \} $, where each session $S_k \in \mathbb{S}_K $ contains $L$ frames with overlap. 
Each session is independently processed by VGGT, producing a set of colored point cloud sessions
$\mathbb{P}_K = \{\mathcal{P}_1, \mathcal{P}_2,...,\mathcal{P}_K \}$,
where $\mathcal{P}_k \in \mathbb{P}_K$ represents the map of sequence $S_k$ aligned to the coordinate frame of its first image.

Our goal is to register each session $\mathcal{P}_k$ with the corresponding LiDAR point cloud $\mathcal{L}_k$ from LIO, and then merge all sessions into a consistent global map. 
Since $\mathcal{P}_k$ lacks an absolute metric scale, while $\mathcal{L}_k$ provides a scaled reference, we estimate a similarity transformation $\mathcal{T}^{V\rightarrow L}_k \in Sim(3)$, which includes a $SE(3)$ transformation and a scale $s \in \mathbb{R}^+$. This $Sim(3)$ transformation aligns each VGGT session with its LiDAR counterpart. 
Next, the relative transformations $T_{k \rightarrow k+1} \in SE(3)$ between adjacent VGGT sessions are refined to achieve seamless fusion and a globally consistent reconstruction.
Our proposed LiDAR-VGGT tackles these challenges, illustrated in Fig.~\ref{fig:system}. 
It takes multiple sequences collected in the same scene, including LiDAR scans, IMU measurements, and images as input, and outputs a dense and globally consistent colored map with metric scale. 
Our system consists of two modules: the Coarse Pre-Fusion Module (Sec.~\ref{subsetcion:frontend}), and the Fine Post-Fusion Module (Sec.~\ref{subsection:backend}).

\subsection{Coarse Pre-Fusion Module}
\label{subsetcion:frontend}

For the VGGT component, each image sequence is downsampled to ensure sufficient disparity and independently processed to estimate camera poses and reconstruct a dense colored map. The resulting poses and point clouds preserve local geometric structures but lack a consistent real-world scale.
For the LIO component, LiDAR and IMU measurements are fused using FAST-LIO2~\cite{fastlio2} to build a LiDAR point cloud map. To ensure long-term global consistency, RING++~\cite{ring++}, a fast LiDAR scan descriptor, is employed for loop closure detection. The detected loop constraints are incorporated into a pose graph, which is subsequently optimized through PGO to obtain a globally consistent LiDAR map.

\textbf{Pose Registration: }
We estimate a $Sim(3)$ transformation for each session $S_k \in \mathbb{S}_K$ using the LiDAR poses to align the VGGT poses $\{T_i^{\text{VGGT}}\}_{i = 1}^L$ to the world coordinate frame. For brevity, the session index $k$ is omitted below.
Each LiDAR pose is converted into the camera pose in the world system $\{T_i^{\text{Cam}}\}_{i = 1}^L$ with the same timestamp using extrinsic parameters. Each VGGT pose is then paired with the nearest camera pose in time: 
\begin{equation}
\begin{aligned}
    \mathcal{C}_k &= \big\{ (T_i^{\text{VGGT}},\, T_{j^*(i)}^{\text{Cam}}) \;\big|\; i = 1,\dots,L \big\}, \\
    j^*(i) &= \arg\min_j \big| t_i^{\text{VGGT}} - t_j^{\text{Cam}} \big|,
\end{aligned}
\end{equation}
where $t_i^{\text{VGGT}}$ and $t_j^{\text{Cam}}$ are the timestamp of the two poses.
Given the poses in pairs, we denote that $\{\mathbf{x}_i\}_{i = 0}^L$ and $\{\mathbf{y}_i\}_{i = 0}^L$ are the translations and rotations of VGGT poses and camera poses. We aim to estimate a $Sim(3)$ transformation $(s_1, \mathbf{R}_1, \mathbf{t}_1)$ including a scale factor and an $SE(3)$ transformation by:
\begin{equation}
    \min_{s_1 > 0,\, \mathbf{R}_1 \in SO(3),\, \mathbf{t}_1 \in \mathbb{R}^3}  
    \sum_{i=1}^{L} \big\| \mathbf{y}_i - \big(s_1 \mathbf{R}_1 \mathbf{x}_i + \mathbf{t}_1 \big) \big\|^2 , 
\end{equation}
which is efficiently solved using the Umeyama method~\cite{umeyama}.

\textbf{Linearity Validation and Rotation Correction: }
Although the Umeyama method efficiently estimates $Sim(3)$ from translations, it may yield inaccurate rotations when trajectories are near-linear.  
We therefore introduce a two-step strategy: linearity validation to detect degenerate motion and rotation correction to refine $\mathbf{R}_1$. 
We perform Principal Component Analysis (PCA) on the camera pose sequence to obtain eigenvalues $\lambda_1 \geq \lambda_2 \geq \lambda_3$ and define linearity as:
\begin{equation}
    \ell = 1 - \frac{\lambda_2 + \lambda_3}{\lambda_1}.
\end{equation}
A large $\ell$ indicates a near-linear trajectory, implying insufficient rotational constraints.  
For such cases, we refine $\mathbf{R}_1$ by aggregating relative rotations between matched VGGT and camera poses.
Let $\{\mathbf{R}_i^{\mathrm{src}}\}_{i=1}^{L}$ denote the source rotations from VGGT, and 
$\{\mathbf{R}_i^{\mathrm{tgt}}\}_{i=1}^{L}$ the target rotations from camera of $L$ frames. We first transform the source rotations by 
\begin{equation}
    \tilde{\mathbf{R}}_i = \mathbf{R}_1\,\mathbf{R}_i^{\mathrm{src}},
\end{equation}
and then calculate the average relative rotation using 
\begin{equation}
    \bar{\mathbf{M}} = \frac{1}{L}\sum_{i=1}^L \mathbf{R}_i^{\mathrm{tgt}}\,\tilde{\mathbf{R}}_i^\top, \qquad i=1,\dots,L.
\end{equation}
To ensure a valid rotation, $\bar{\mathbf{M}}$ is projected onto $SO(3)$ via SVD. $\bar{\mathbf{M}}=\mathbf{U}\mathbf{\Sigma}\mathbf{V}^\top$ is the singular value decomposition:
\begin{equation}
    \Delta\mathbf{R} \;=\; \mathbf{U}\,\mathrm{diag}(1,1,\det(\mathbf{U}\mathbf{V}^\top))\,\mathbf{V}^\top,
\end{equation}
and the corrected rotation is:
\begin{equation}
    \mathbf{R}_1^{*} \;=\; \Delta\mathbf{R}\,\mathbf{R}_1.
\end{equation}

\textbf{Scale RANSAC and Refinement: }
\begin{figure}[!t]
  \centering
  \includegraphics[width=1.0\linewidth]{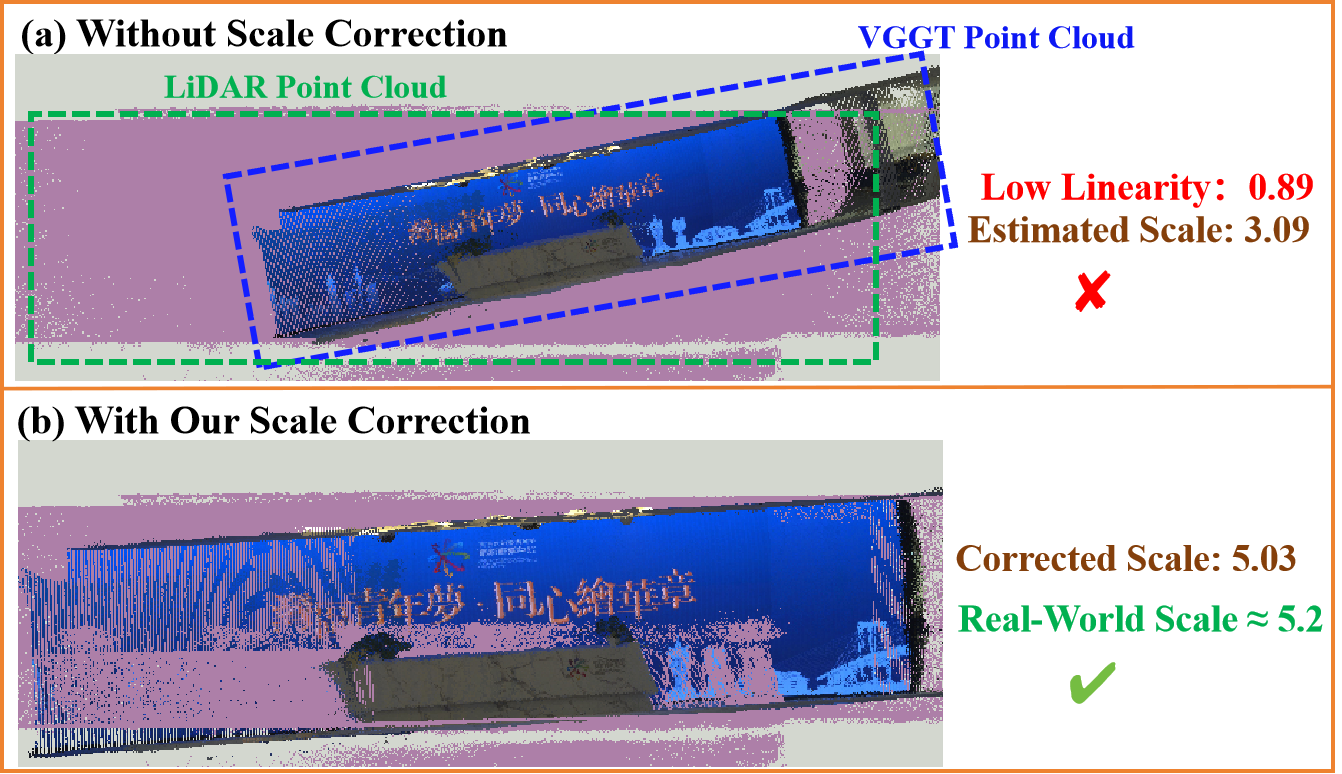}
   \vspace{-20pt}
  \caption{(a) demonstrates an example with low linearity of LIO poses and thus yielding incorrect scale after pose registration. (b) shows the result with our scale correction which recovers an accurate scale close to the real value.}
  \label{fig:linearity}
\end{figure}
Inaccurate VGGT poses can distort scale estimation , as shown in Fig.~\ref{fig:linearity}(a). We thus employ a RANSAC-based approach to identify reliable scales and refine session-level scaling.  
In particular, pose sequences with high linearity tend to yield more accurate scale estimates, as they benefit from robust pose estimation and well-aligned correspondences. In contrast, sequences with low linearity usually involve aggressive motions, making it difficult to recover the correct scale. Therefore, we prioritize sequences with high linearity as inlier priors in our scale RANSAC.  

Let $\{s_k\}_{k=1}^K$ denote initial scale estimates and $\{\ell_k\}_{k=1}^K$ their linearity scores.  
We define a scale threshold based on the standard deviation:  
\begin{equation}
    \mathrm{threshold} = k_\sigma \cdot \sigma_s, \quad \sigma_s = \mathrm{std}(\{s_k\}),
\end{equation}
where $k_\sigma$ is set to 2. High-linearity frames are prioritized in RANSAC by converting $\ell_k$ to sampling probabilities via softmax:
\begin{equation}
    p_k = \frac{\exp(\alpha \ell_k)}{\sum_{k=1}^{K} \exp(\alpha \ell_k)},
\end{equation}
where $\alpha$ is calculated by the coefficient of variation $cov$: 
\begin{equation}
    \alpha = \frac{1}{cov} =  \frac{\frac{1}{K} \sum_{k = 1}^{K} s_k}{\sigma_s}.
\end{equation}
For a candidate scale $s_c$, inliers are defined as:
\begin{equation}
    \mathcal{I}(s_c) = \{k \mid |s_k - s_c| < \mathrm{threshold} \}.
\end{equation}
After iterations, scale outliers are corrected by combining the nearest inlier scale with the $Sim(3)$ estimated scale from their overlapping poses, thus ensuring consistent and correct scale propagation, as shown in Fig.~\ref{fig:linearity}(b).
\subsection{Fine Post-Fusion Module}
\label{subsection:backend}

The post-fusion module further refines coarse alignment through enhanced cross-modal $Sim(3)$ registration and global PGO, ensuring metric-scale accuracy and global consistency in large-scale reconstruction.

\textbf{Enhanced Cross-Modal Sim(3) Registration: }
To further register the VGGT point cloud session $\mathcal{P}_k$ to the corresponding LiDAR session $\mathcal{L}_k$ and obtain the real-world scale, we first transform $\mathcal{P}_k$ using the initial estimate in Sec.~\ref{subsetcion:frontend} to obtain $\mathcal{P}'_k$. This provides an initial value for subsequent registration. 
However, due to the field-of-view (FOV) discrepancy between camera and LiDAR, unconstrained $Sim(3)$ optimization may overfit and cause scale drift.  
To stabilize scale, we introduce a regularized objective:
\begin{equation}
\text{%
        \small
    $\min_{s_2 > 0,\, \mathbf{R}_2 \in SO(3),\, \mathbf{t}_2 \in \mathbb{R}^3} 
    \sum_{i=1}^M \big\| \mathbf{q}_i - (s_2 \mathbf{R}_2 \mathbf{p}_i + \mathbf{t}_2 ) \big\|^2 + \lambda (s_2 - s_1^*)^2,$
    }
\end{equation}
where the regularization weight $\lambda$ is defined as:
\begin{equation}
\lambda = \beta \cdot n \cdot D^2,  \quad
\beta \in (0, 1],
\end{equation}
and $n$ denotes the number of points in the source map, $D$ represents the diagonal length of the bounding box, and $\beta$ is a tunable coefficient controlling the strength of the regularization.

The regularized optimization is solved through an alternating strategy: fixing $\mathbf{R}_2,\mathbf{t}_2$ to update $s_2$, followed by refining $\mathbf{R}_2,\mathbf{t}2$ based on nearest-neighbor correspondences. The closed-form solution for $s_2$ is:
\begin{equation}
s_2 = \frac{\sum_{i=1}^M (\mathbf{q}_i - \mathbf{t}_2)^\top \mathbf{R}_2 \mathbf{p}_i + \lambda s_1^*}
{\sum_{i=1}^M \big\| \mathbf{R}_2 \mathbf{p}_i \big\|^2 + \lambda}.
\end{equation}
The regularization term prevents the scale from drifting excessively from the initial estimate $s_1^*$, ensuring stable cross-modal alignment. 

\begin{figure}[!t]
  \centering
  \includegraphics[width=0.96\linewidth]{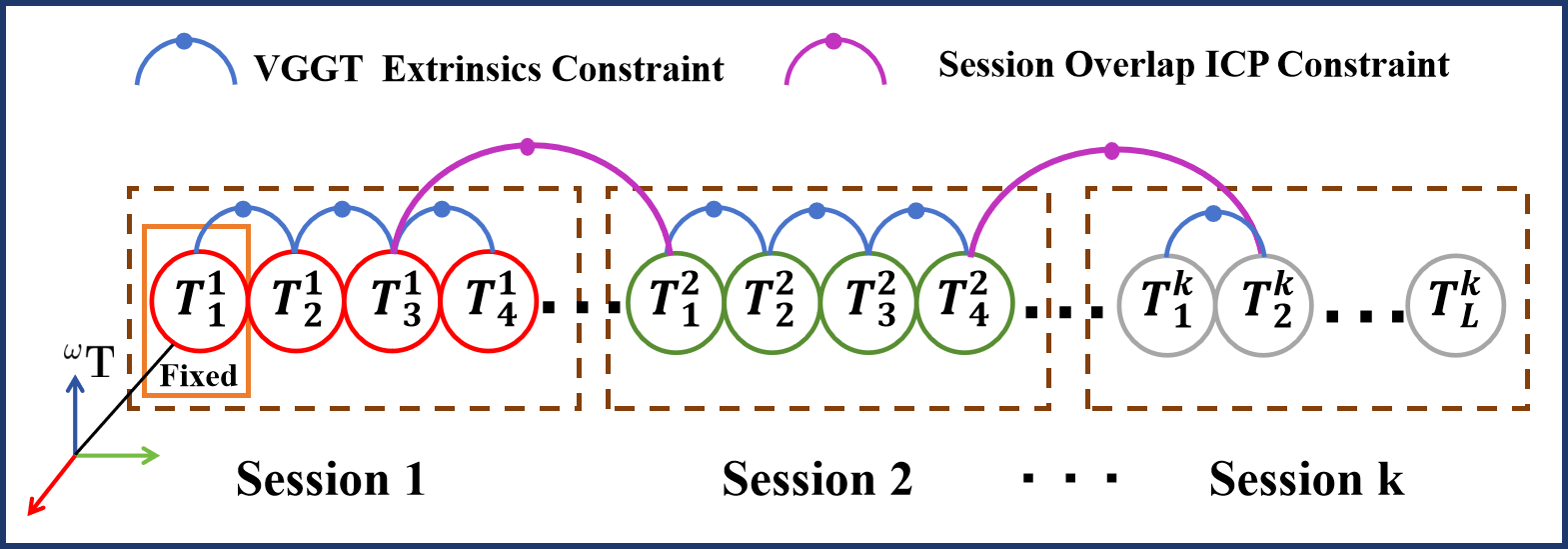}
   \vspace{-10pt}
  \caption{Structure of our global PGO. The first frame of the first session is fixed as the world reference frame.}
  \label{fig:pgo}
   \vspace{-0.3cm}
\end{figure}

\textbf{Global Pose Graph Optimization: }
While session-level registration ensures local accuracy, global consistency is achieved through PGO.
Fig.~\ref{fig:pgo} illustrates our framework. It includes two types of constraints: (i) intra-session constraints from VGGT poses, and (ii) inter-session constraints obtained by applying ICP on overlapping regions. We implement this optimization in GTSAM~\cite{gtsam}, and the refined poses are then propagated to the colored point clouds of all frames, resulting in a globally consistent reconstruction.

\subsection{Colored Map Evaluation}
We propose four metrics to evaluate colored point cloud quality: color distance (CD), color fidelity (CF), local color recall (LCR), and color consistency score (CCS).  

\textbf{Color Distance and Color Fidelity: }
The CD measures the average bidirectional color difference between the reconstructed and reference point clouds, and CF expresses it in decibel form for interpretability.
\begin{align}
\mathrm{CD} &= \frac{1}{2N_s} \sum_{i=1}^{N_s} \| c_i^s - c_{\pi(i)}^r \|_2
+ \frac{1}{2N_r} \sum_{j=1}^{N_r} \| c_j^r - c_{\pi'(j)}^s \|_2, \nonumber \\
\mathrm{CF} &= -20 \log_{10}(\mathrm{CD}),
\end{align}
where $c_i^s$ and $c_j^r$ denote the RGB color vectors of the source and reference point clouds, 
$N_s$ and $N_r$ are their point numbers, and 
$\pi(i)$ and $\pi'(j)$ represent the nearest-neighbor correspondences between the two point sets.
Smaller $\mathrm{CD}$ and higher $\mathrm{CF}$ indicate better color fidelity.

\textbf{Local Color Recall: }
The LCR measures the proportion of reference points whose local color difference in the reconstructed map is below a threshold:
\begin{equation}
    \mathrm{LCR}_{r \to s}(\tau, r_g) = \frac{1}{N_r} \sum_{j=1}^{N_r} \mathbb{I} \left( \min_{i \in \mathcal{N}_s(x_j^r; r_g)} \| c_i^s - c_j^r \|_2 \le 3\tau \right),
\end{equation}
where $\mathcal{N}_s(x_j^r; r_g)$ is the neighborhood of $x_j^r$ within radius $r_g$, $\tau$ is the color threshold, and $\mathbb{I}(\cdot)$ is the indicator function which returns $1$ if the condition holds and $0$ otherwise.

\textbf{Color Consistency Score: }
We assume that points within a small voxel of a colored map have similar RGB values due to spatial continuity. The CCS quantifies this local color homogeneity by averaging the RGB covariance trace of all voxels:
\begin{equation}
\begin{aligned}
\mathrm{CCS} &= \frac{1}{N_v} \sum_{k=1}^{N_v} C_i^{(k)}, \quad
C_i^{(k)} = \mathrm{tr}(\Sigma_c^{(k)}), \\
\Sigma_c^{(k)} &= \frac{1}{n_k - 1} \sum_{j=1}^{n_k} (c_j^{(k)} - \bar{c}^{(k)})(c_j^{(k)} - \bar{c}^{(k)})^\top, 
\end{aligned}
\end{equation}
where $N_v$ is the number of occupied voxels, $n_k$ is the number of points in the $k$-th voxel, 
$c_j^{(k)}$ and $\bar{c}^{(k)}$ are the RGB vector and its mean in voxel $k$, and 
$\Sigma_c^{(k)}$ is the RGB covariance matrix. 
Smaller $\mathrm{CCS}$ indicates more consistent colors.

\section{Experiments}
\subsection{Experimental Setup}

\begin{figure}[!t]
  \centering
  \includegraphics[width=1.0\linewidth]{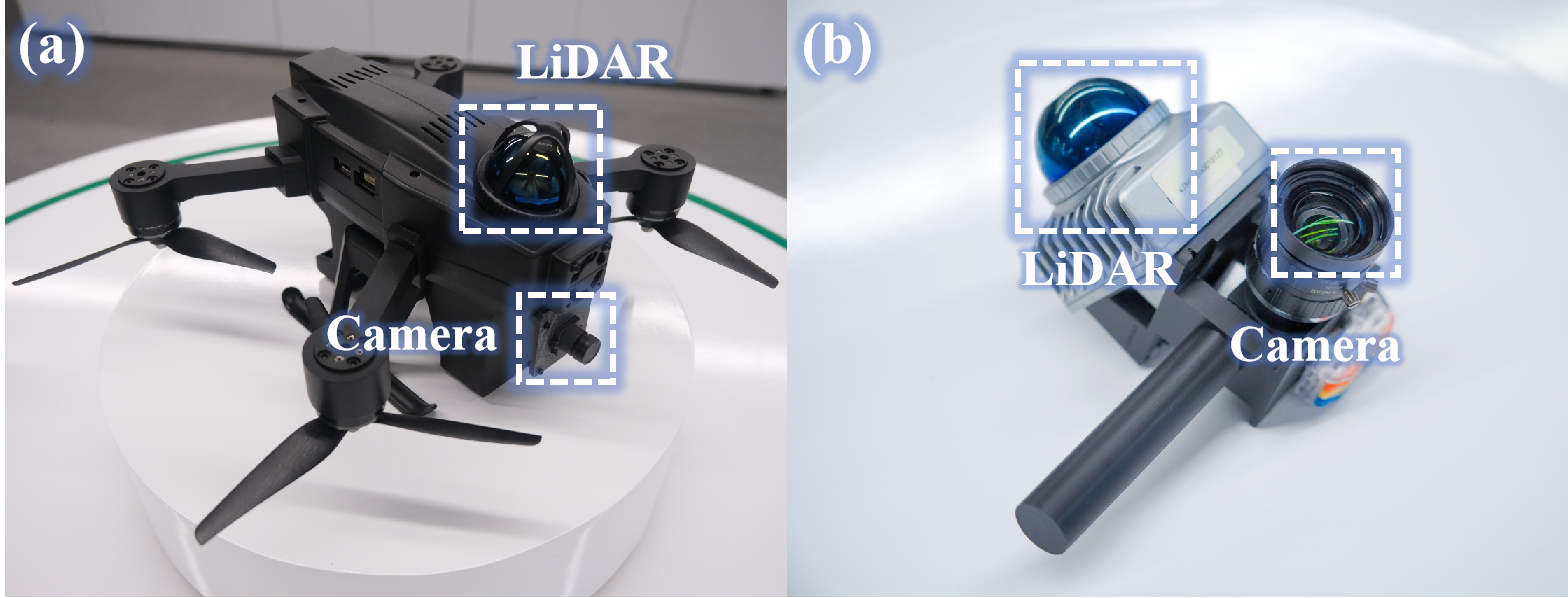}
  \vspace{-20pt}  
  \caption{Our two data collection platforms. The UAV device (a) is used for outdoor data collection, while the handheld device (b) is used for indoor.}
  \label{fig:registration}
\end{figure}

\begin{table}[t]
  \centering
  \caption{Overview of Dataset for Experiments}
  \vspace{-0.2cm}
  \label{table:dataset}
  \begin{threeparttable}
  \begin{tabular*}{0.48\textwidth}{@{  } @{\extracolsep{\fill}} c c c c @{  }}
    \toprule
    \makecell{\textbf{Dataset}} & 
    \makecell{\textbf{Sequence}} &
    \makecell{\textbf{Trajectory} \\ \textbf{Length (m)}} & 
    \makecell{\textbf{LiDAR}} \\
    \midrule
    \multirow{4}{*}{\textbf{MARS-LVIG}}
        & HKAirport01 & 2058 & Avia\\
        & HKisland01 & 1877 & Avia\\
        & AMvalley01 & 4377 & Avia \\
        & AMtown01 & 4915 & Avia \\
        & HKairport03 & 2073 & Avia\\
    \midrule
    \multirow{3}{*}{\textbf{FAST-LIVO2}}
        & Brightwall & 24 & Avia \\
        & CBDBuilding & 34 & Avia \\
    \midrule
    \multirow{3}{*}{\textbf{MUN\_FRL}}
        & LightHouse & 168 & Velodyne  \\
        & Flight & 319 & Velodyne \\
    \midrule
    \multirow{3}{*}{\textbf{Self-Collect}}
        & Company & 75 & MID360 \\
        & Street & 69 & MID360 \\
        & TechnologyPark  & 45 & MID360 \\
        \bottomrule
  \end{tabular*}
  \end{threeparttable}
  \vspace{-0.3cm}
\end{table}

To evaluate the performance of LiDAR-VGGT, we conduct experiments on several datasets, including the public MARS-LVIG \cite{lvig}, FAST-LIVO2 \cite{fastlivo2} and MUN\_FRL \cite{munfrl} dataset, and a self-collect dataset. and their parameters are shown in Tab.~\ref{table:dataset}. 
We first assess the overall mapping quality, benchmarking our approach against VGGT-Long \cite{vggt-long}, VGGT-SLAM \cite{vggt-slam} and SLAM3R \cite{slam3r}. The evaluation was carried out from two complementary perspectives: geometric fidelity and color quality.
We then perform a registration experiment for ablation to evaluate the individual contributions of our pre-fusion and post-fusion modules, with particular attention to the impact of the proposed enhanced $Sim(3)$ with regularization.
Finally, a robustness and sensitivity analysis is carried out to demonstrate that our pipeline exhibits greater resilience to extrinsic noise compared to state-of-the-art LIVO methods FAST-LIVO2.

\subsection{Mapping Quality Experiment}

\begin{figure}[!t]
  \centering
  \includegraphics[width=1.0\linewidth]{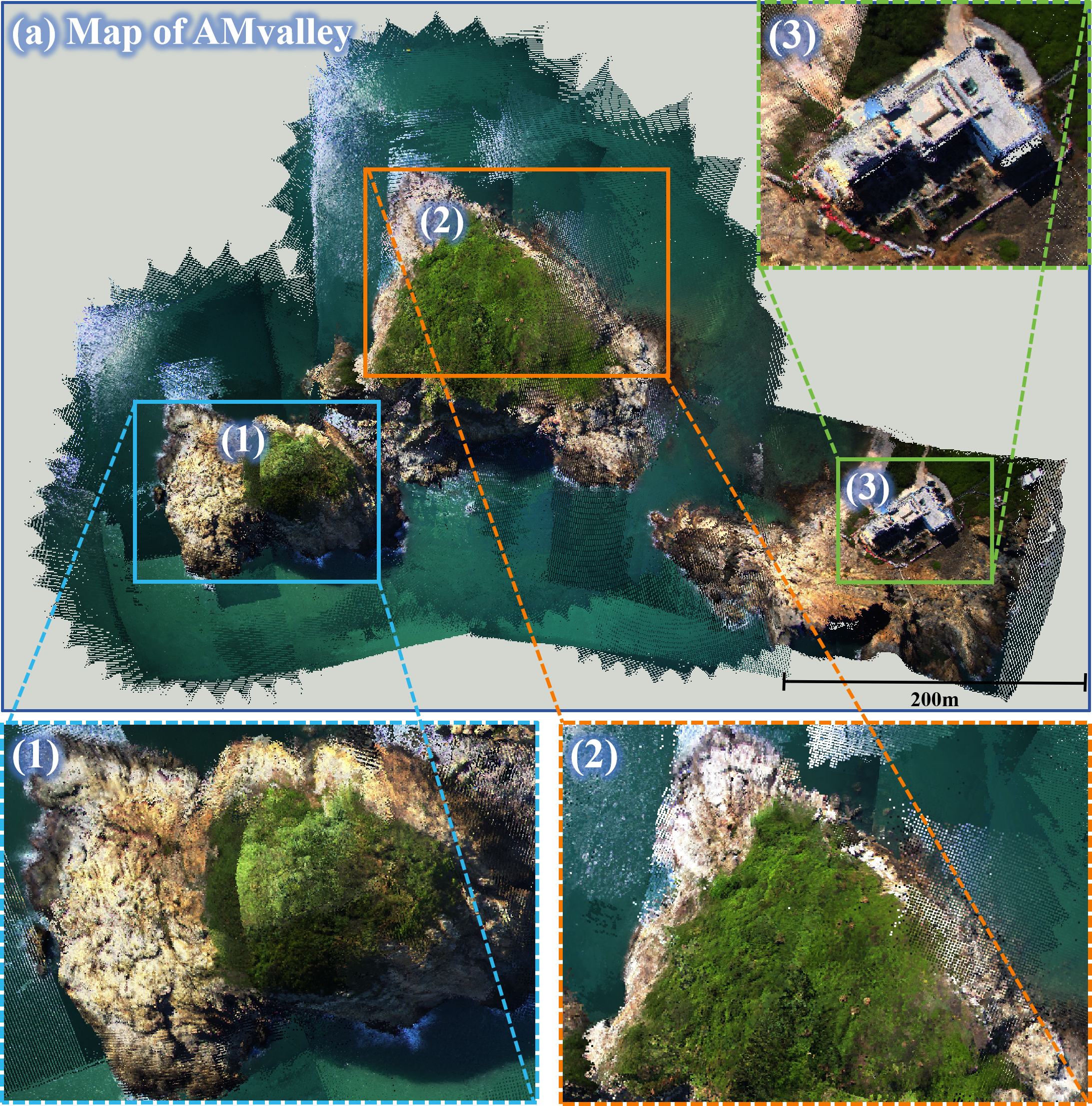}
  \vspace{-20pt}  
  \caption{Our large-scale mapping result in the MARS-LVIG HKisland01 dataset. The enlarged details show that the reconstructed point cloud is dense, consistent, and maintains an accurate metric scale.}
  \label{fig:mapping}
\end{figure}

\begin{table}[t]
\centering
\caption{Mapping Geometric Fidelity Comparison}
\label{tab:multimappinggeo}
\renewcommand{\arraystretch}{1.0}
\setlength{\tabcolsep}{3pt}
\begin{threeparttable}
\begin{tabular}{@{}llccccc@{}}
\toprule[0.03cm]
 & \textbf{Data} & \textbf{V-Long} & \textbf{V-Sim(3)} & \textbf{SLAM3R} & \textbf{V-SL(4)} & \textbf{Ours} \\
\midrule[0.03cm]

\multirow{5}{*}{\shortstack[1]{\textbf{CD\#}\\ (m)$\downarrow$}} & AMtown01 & 52.01 & \cellcolor{orange!35}51.34 & 184.4 & $\times$ & \cellcolor{red!35}2.23 \\
 & AMvalley01 & 55.14 & \cellcolor{orange!35}51.12 & 135.4 & $\times$ & \cellcolor{red!35}10.2 \\
 & HKisland01 & 26.2 & \cellcolor{orange!35}17.14 & 35.5 & $\times$ & \cellcolor{red!35}6.33 \\
 & LightHouse & 25.49 & 34.9 & \cellcolor{orange!35}15.5 & 23.25 & \cellcolor{red!35}2.34 \\
 & Flight & 20.09 & 16.92 & 87.9 & \cellcolor{orange!35}14.0 & \cellcolor{red!35}3.7 \\
\midrule

\multirow{5}{*}{\shortstack[1]{\textbf{AC} \\ (m)$\downarrow$}} & AMtown01 & \cellcolor{orange!35}0.044 & \cellcolor{orange!35}0.044 & \cellcolor{red!35}0.04 & $\times$ & \cellcolor{red!35}0.04 \\
 & AMvalley01 & 0.04 & 0.043 & \cellcolor{orange!35}0.036 & $\times$ & \cellcolor{red!35}0.035 \\
 & HKisland01 & \cellcolor{red!35}0.033 & \cellcolor{orange!35}0.035 & \cellcolor{red!35}0.033 & $\times$ & 0.039 \\
 & LightHouse & \cellcolor{orange!35}0.03 & \cellcolor{orange!35}0.03 & 0.036 & \cellcolor{red!35}0.028 & 0.04 \\
 & Flight & 0.044 & 0.042 & \cellcolor{red!35}0.04 & \cellcolor{orange!35}0.041 & \cellcolor{red!35}0.04 \\
\midrule

\multirow{5}{*}{\shortstack[1]{\textbf{ICP} \\ \textbf{Overlap} \\ $\uparrow$}} & AMtown01 & 8\% & 1.2\% & \cellcolor{orange!35}9.3\% & $\times$ & \cellcolor{red!35}53\% \\
 & AMvalley01 & 2.3\% & 2.2\% & \cellcolor{orange!35}4.9\% & $\times$ & \cellcolor{red!35}36.8\% \\
 & HKisland01 & 13.5\% & \cellcolor{orange!35}28.2\% & 21.6\% & $\times$ & \cellcolor{red!35}58.6\% \\
 & LightHouse & 8.8\% & 16.5\% & \cellcolor{orange!35}32.5\% & 12.9\% & \cellcolor{red!35}91\% \\
 & Flight & \cellcolor{orange!35}33\% & 23.5\% & 8.3\% & 22.4\% & \cellcolor{red!35}78.9\% \\
\midrule

\multirow{5}{*}{\shortstack[1]{\textbf{AWD} \\ (m)$\downarrow$}} & AMtown01 & 1.67 & 1.4 & \cellcolor{orange!35}1.37 & $\times$ & \cellcolor{red!35}0.8 \\
 & AMvalley01 & 1.89 & \cellcolor{orange!35}1.75 & 1.79 & $\times$ & \cellcolor{red!35}1.59 \\
 & HKisland01 & 1.64 & \cellcolor{orange!35}1.34 & 1.42 & $\times$ & \cellcolor{red!35}1.12 \\
 & LightHouse & 1.88 & \cellcolor{orange!35}1.67 & 1.92 & 1.80 & \cellcolor{red!35}1.41 \\
 & Flight & \cellcolor{orange!35}1.59 & 1.66 & 1.69 & 1.78 & \cellcolor{red!35}1.50 \\
\midrule

\multirow{5}{*}{\shortstack[1]{\textbf{Fitness} \\ $\uparrow$}} & AMtown01 & \cellcolor{orange!35}0.007 & 0.001 & 0.004 & $\times$ & \cellcolor{red!35}0.05 \\
 & AMvalley01 & 0.0012 & 0.002 & \cellcolor{orange!35}0.003 & $\times$ & \cellcolor{red!35}0.023 \\
 & HKisland01 & 0.006 & \cellcolor{orange!35}0.015 & 0.01 & $\times$ & \cellcolor{red!35}0.04 \\
 & LightHouse & 0.0043 & 0.009 & \cellcolor{orange!35}0.03 & 0.0064 & \cellcolor{red!35}0.13 \\
 & Flight & \cellcolor{orange!35}0.036 & 0.022 & 0.006 & 0.022 & \cellcolor{red!35}0.08 \\
\bottomrule[0.03cm]
\end{tabular}
\begin{tablenotes}
\footnotesize
\item \# CD denotes Chamfer distance, $\times$ denotes system fails.
\end{tablenotes}
\end{threeparttable}
\vspace{-0.5em}
\end{table}


\begin{table}[t]
\centering
\caption{Colored Point Cloud Evaluation}
\vspace{-0.2cm}
\label{tab:multimappingcolor}
\renewcommand{\arraystretch}{1.0}
\setlength{\tabcolsep}{5pt}
\begin{threeparttable}
\begin{tabular}{@{}llcccc@{}}
\toprule[0.03cm]
 & \textbf{Data} & \textbf{V-Long} & \textbf{V-Sim(3)} & \textbf{SLAM3R} & \textbf{Ours} \\
\midrule[0.03cm]

\multirow{4}{*}{\shortstack[1]{\textbf{CD*} $\downarrow$}} 
 & AMtown01 & 130.49 & 130.96 & \cellcolor{orange!35}122.22 & \cellcolor{red!35}82.93 \\
 & AMvalley01 & 87.04 & 85.76 & \cellcolor{orange!35}85.50 & \cellcolor{red!35}84.48 \\
 & HKisland01 & 158.21 & \cellcolor{orange!35}142.29 & 152.83 & \cellcolor{red!35}112.90 \\
 & LightHouse & \cellcolor{orange!35}67.92 & 70.80 & 72.23 & \cellcolor{red!35}58.56 \\
 & Flight & 59.84 & \cellcolor{orange!35}57.98 & 59.23 & \cellcolor{red!35}57.55 \\
\midrule

\multirow{5}{*}{\shortstack[1]{\textbf{CF} (dB)$\uparrow$}} 
 & AMtown01 & 11.87 & 11.84 & \cellcolor{orange!35}12.41 & \cellcolor{red!35}15.81 \\
 & AMvalley01 & 15.39 & 15.52 & \cellcolor{orange!35}15.54 & \cellcolor{red!35}15.57 \\
 & HKisland01 & 10.21 & \cellcolor{orange!35}11.09 & 10.50 & \cellcolor{red!35}13.12 \\
 & LightHouse & \cellcolor{orange!35}11.49 & 11.13 & 10.96 & \cellcolor{red!35}12.78 \\
 & Flight & 12.59 & \cellcolor{orange!35}12.87 & 12.68 & \cellcolor{red!35}12.93 \\
\midrule

\multirow{5}{*}{\shortstack[1]{\textbf{CCS} $\downarrow$}} 
 & AMtown01 & \cellcolor{orange!35}1.53 & 2.07 & 1.76 & \cellcolor{red!35}1.19 \\
 & AMvalley01 & 0.83 & 0.88 & \cellcolor{orange!35}0.73 & \cellcolor{red!35}0.67 \\
 & HKisland01 & 1.99 & \cellcolor{orange!35}1.76 & 1.81 & \cellcolor{red!35}1.11 \\
 & LightHouse & \cellcolor{red!35}1.54 & 2.18 & 1.72 & \cellcolor{orange!35}1.62 \\
 & Flight & 1.31 & 1.75 & \cellcolor{orange!35}1.27 & \cellcolor{red!35}1.12 \\
\midrule

\multirow{5}{*}{\shortstack[1]{\textbf{LCR} $\uparrow$}} 
 & AMtown01 & \cellcolor{orange!35}2.6\% & 0.4\% & 0.86\% & \cellcolor{red!35}24.1\% \\
 & AMvalley01 & \cellcolor{orange!35}1.32\% & 1.27\% & 0.5\% & \cellcolor{red!35}17.8\% \\
 & HKisland01 & 3.2\% & \cellcolor{orange!35}5.96\% & 2.8\% & \cellcolor{red!35}23.0\% \\
 & LightHouse & 1.7\% & 4.6\% & \cellcolor{orange!35}6.1\% & \cellcolor{red!35}36.7\% \\
 & Flight & \cellcolor{orange!35}7.9\% & 3.1\% & 0.8\% & \cellcolor{red!35}18.6\% \\
\bottomrule[0.03cm]
\end{tabular}
\begin{tablenotes}
\footnotesize
\item * CD denotes color distance.
\end{tablenotes}
\end{threeparttable}
\vspace{-0.5em}
\end{table}

This section evaluates the overall mapping quality of our method, VGGT-Long, VGGT-SLAM, and SLAM3R. For simplicity, we refer to VGGT-Long as V-Long, and the two modes of VGGT-SLAM as V-Sim(3) and V-SL(4), respectively. We use the map evaluation tool~\cite{mapeval} and our proposed tool to analyze geometric fidelity and color quality against ground-truth colored point cloud maps. For MARS-LVIG, the ground-truth map is a high-precision point cloud generated by the DJI L1 and DJI Terra system. For MUN\_FRL, it is produced by interpolating ground-truth poses with known extrinsics and projecting images onto the point cloud. For benchmarking, we use CloudCompare to match the scale with the ground-truth map and align the first frame.

The results of the geometric fidelity experiment are presented in Tab.~\ref{tab:multimappinggeo}. Our method achieves the best performance on all four metrics, including Chamfer distance (CD), ICP overlap ratio, average Wasserstein distance (AWD), and Fitness. Except for the most challenging AMvalley01 dataset, our approach attains an overlap ratio exceeding 50\% on the remaining four sequences, demonstrating that it effectively preserves both the true scale and geometric structure of the scene. The large-scale datasets cover hundreds to thousands of meters. In such scenarios, the LiDAR SLAM alone accumulates noticeable mapping errors. However, LiDAR-VGGT achieves an AWD of nearly one meter, which demonstrates that our point cloud maintains high geometric consistency, closely matching the ground truth. Moreover, our method achieves an average CD of only a few meters. The slightly higher CD in some cases is mainly due to FOV differences, where the bidirectional calculation amplifies errors in non-overlapping regions. Finally, our method achieves the best accuracy distance (AC) on three datasets. As AC depends on the number of inlier point pairs, benchmark methods with partial map overlap naturally yield fewer inliers and lower AC, which is reasonable.

\begin{figure}[!t]
  \centering
  \vspace{-3pt}
  \includegraphics[width=1.0\linewidth]{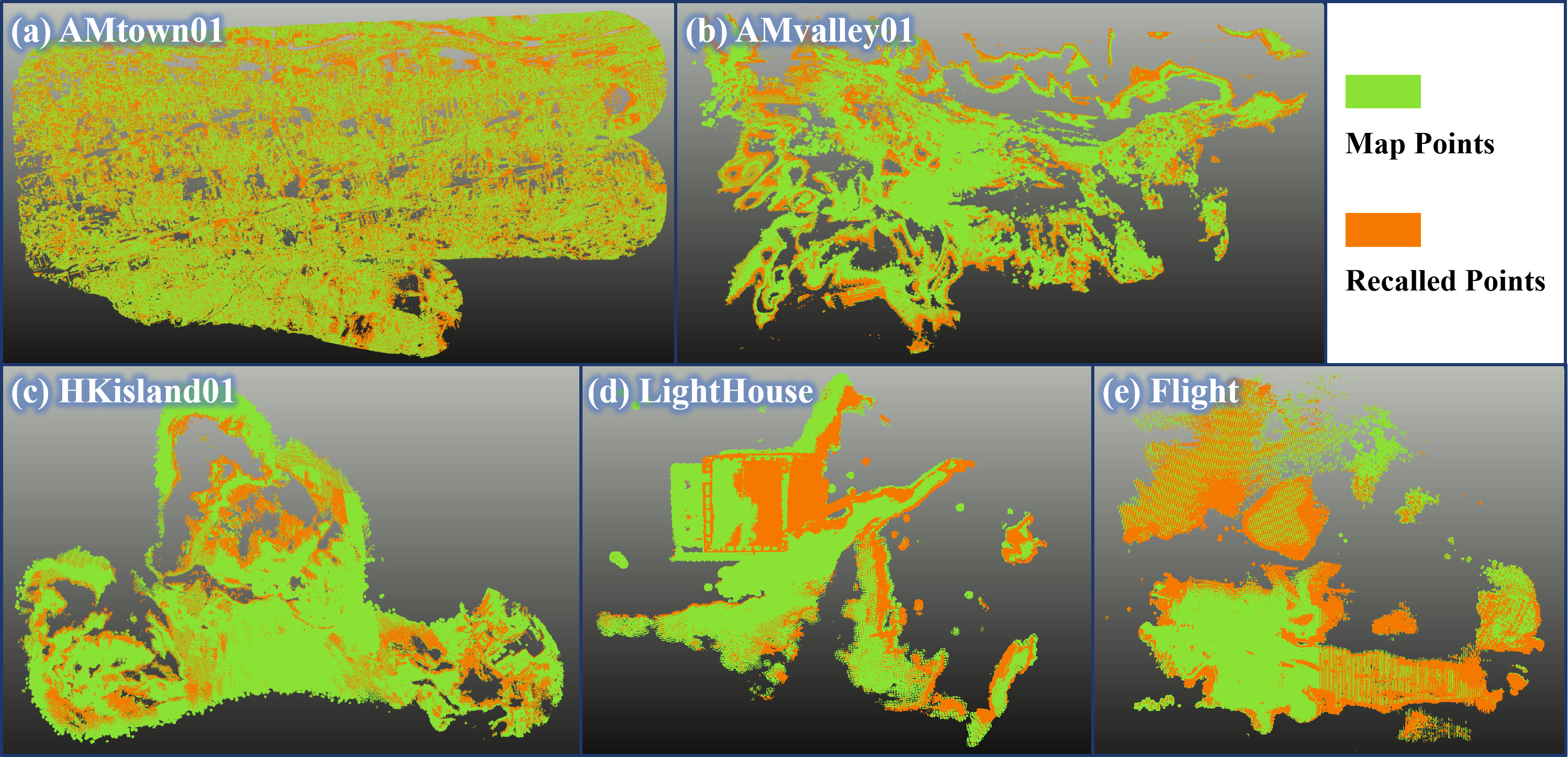}
   \vspace{-20pt}
  \caption{The visualization of LCR on five sequences shows that our mapping results maintain good local color consistency compared to the ground-truth map. The holes in the middle are mainly caused by the lower geometric quality of the VGGT point clouds, which deviate from the truth map.}
  \label{fig:recall}
  \vspace{-0.3cm}
\end{figure}

The color quality evaluation results are shown in Tab.~\ref{tab:multimappingcolor}. Our method achieves near-optimal performance on all metrics, demonstrating excellent color fidelity. Fig.~\ref{fig:recall} shows the LCR at $\tau=0.1$ and $r_g=0.5m$ in all five datasets, illustrating that our map maintains overall color consistency with the ground-truth, while the empty areas mainly result from the poor geometric quality of the VGGT point cloud with little overlap in this region.
Since the benchmark methods almost fail on the challenging dataset, only our mapping results are shown in Fig.\ref{fig:mapping}. As seen, our method produces high-quality, dense maps over a large-scale island scene.

\subsection{Registration Experiment for Ablation}

We use the LiDAR point cloud as the reference to evaluate registration performance and compare three variants: the normal $Sim(3)$, our proposed enhanced cross-modal $Sim(3)$, and the pre-fusion module with only pose registration. We compute root mean square error (RMSE), Chamfer distance (CD), average Wasserstein distance (AWD), and spatial consistency score (SCS) for each session and report their averages for all sessions using Map\_Eval~\cite{mapeval}. Tab.~\ref{tab:registration_compact} 
presents a comparison of three registration approaches, two of which use the same number of iterations. Our method consistently outperforms the normal $Sim(3)$ across nearly all metrics, whereas the pre-fusion approach performs the worst due to the lack of direct point cloud fusion. Fig.~\ref{fig:registration} provides a direct explanation for the superior performance of our method. In cross-modal matching with differing point cloud FOVs, unconstrained scale updates in normal $Sim(3)$ cause unbounded scale distortion, as seen in (b) and (d). In contrast, the proper incorporation of the regularization term in our method effectively preserves an accurate scale as demonstrated in Fig.~\ref{fig:registration} (a), (c), and (e).

\begin{figure}[!t]
  \centering
  \includegraphics[width=1.0\linewidth]{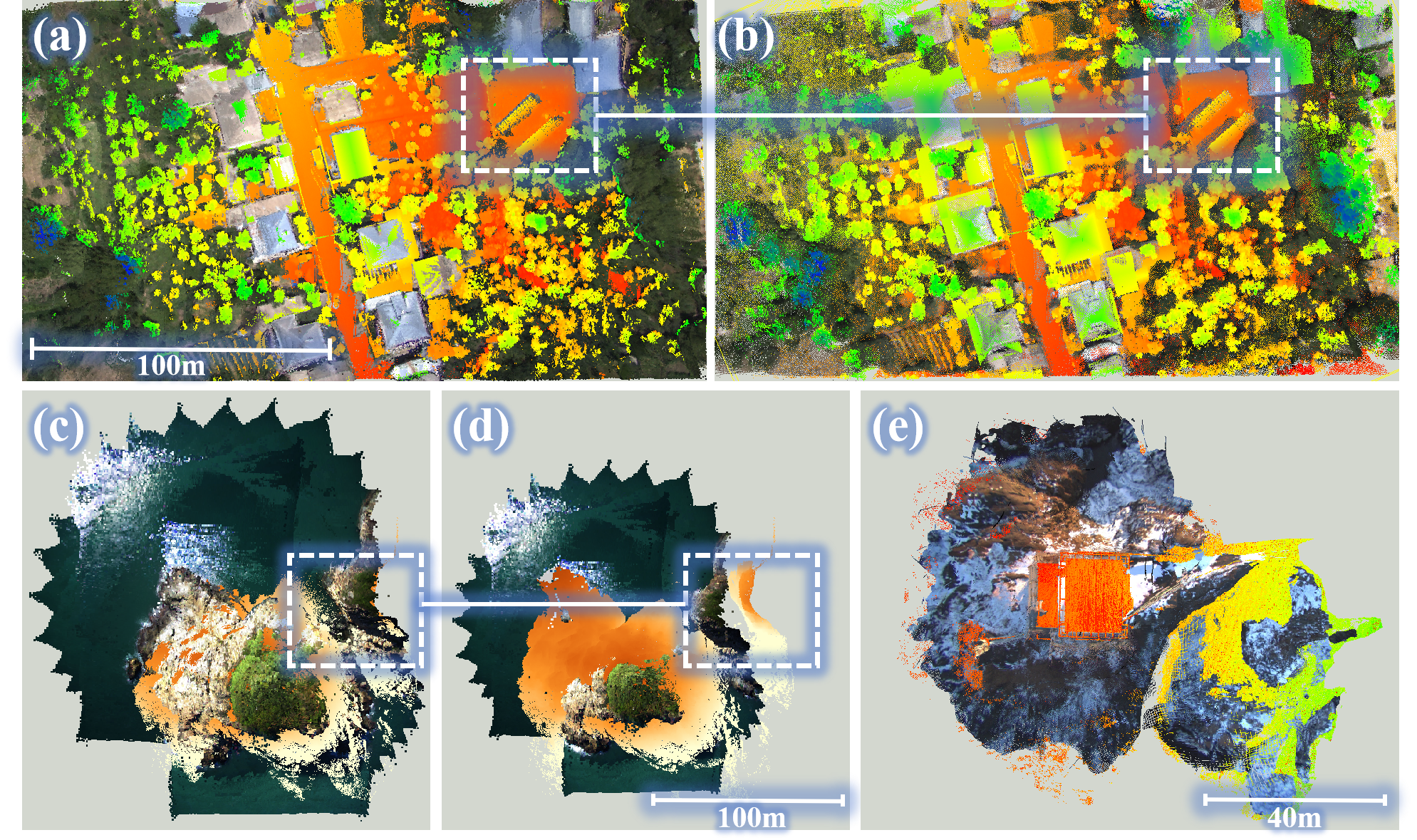}
   \vspace{-20pt}
  \caption{These figures show registration comparisons on a single session of the MARS-LVIG dataset between normal $Sim(3)$ and our regularized $Sim(3)$. The former suffers from excessive scale distortion, while the latter achieves accurate alignment with minimal scale error. (e) presents the registration result of our method on the MUN\_FRL dataset.}
  \label{fig:registration}
  \vspace{-0.3cm}
\end{figure}

\begin{table*}[t]
\centering
\caption{Registration Comparison (Compact Format)}
\vspace{-0.2cm}
\label{tab:registration_compact}
\renewcommand{\arraystretch}{1.0}
\begin{threeparttable}
\begin{tabular}{@{}l cccc cccc cccc cccc@{}}
\toprule
\multirow{2}{*}{\textbf{Data}} &
\multicolumn{4}{c}{\textbf{Pre-Fusion}} &
\multicolumn{4}{c}{$\mathbf{Normal\text{~}Sim(3)}$} &
\multicolumn{4}{c}{$\mathbf{Regularization\text{~} Sim(3)}$} \\ 
\cmidrule(lr){2-5} \cmidrule(lr){6-9} \cmidrule(lr){10-13}
& \textbf{RMSE (m)$\downarrow$} & \textbf{AWD (m)$\downarrow$} & \textbf{CD (m)$\downarrow$} & \textbf{SCS$\downarrow$}
& \textbf{RMSE$\downarrow$} & \textbf{AWD$\downarrow$} & \textbf{CD$\downarrow$} & \textbf{SCS$\downarrow$}
& \textbf{RMSE$\downarrow$} & \textbf{AWD$\downarrow$} & \textbf{CD$\downarrow$} & \textbf{SCS$\downarrow$} \\

\midrule

HKairport03 & 72.87 & 5.89 & 2.45 & $\times$ & \cellcolor{orange!35}1.03 & \cellcolor{orange!35}1.81 & \cellcolor{orange!35}0.051 & \cellcolor{red!35}0.54 & \cellcolor{red!35}1.02 & \cellcolor{red!35}1.34 & \cellcolor{red!35}0.048 & \cellcolor{orange!35}0.6 \\

HKisland01 & 37.31 & 6.76 & 22.1 & $\times$ & \cellcolor{orange!35}1.01 & \cellcolor{orange!35}0.74 & \cellcolor{orange!35}0.05 & \cellcolor{orange!35}0.6 & \cellcolor{red!35}0.968 & \cellcolor{red!35}0.73 & \cellcolor{red!35}0.046 & \cellcolor{red!35}0.56 \\

AMtown01 & 17.08 & 14.26 & 10.02 & $\times$ & \cellcolor{orange!35}0.901 & \cellcolor{orange!35}0.786 & \cellcolor{orange!35}0.057 & \cellcolor{orange!35}0.53 & \cellcolor{red!35}0.893 & \cellcolor{red!35}0.675 & \cellcolor{red!35}0.056 & \cellcolor{red!35}0.52 \\
HKairport01 & 38.23 & 7.77 & 23.0 & $\times$ & \cellcolor{orange!35}1.03 & \cellcolor{orange!35}1.29 & \cellcolor{orange!35}0.042 & \cellcolor{red!35}0.5 & \cellcolor{red!35}0.97 & \cellcolor{red!35}1.25 & \cellcolor{red!35}0.032 & \cellcolor{red!35}0.5 \\

AMvalley01 & 224.9 & 10.39 & 2.53 & $\times$ & \cellcolor{orange!35}2.95 & \cellcolor{orange!35}2.49 & \cellcolor{orange!35}0.031 & \cellcolor{orange!35}0.66 & \cellcolor{red!35}0.942 & \cellcolor{red!35}1.33 & \cellcolor{red!35}0.018 & \cellcolor{red!35}0.64 \\

BrightWall & 2.06 & 0.88 & 0.101 & 0.61 & \cellcolor{orange!35}0.205 & \cellcolor{orange!35}0.66 & \cellcolor{orange!35}0.09 & \cellcolor{orange!35}0.60 & \cellcolor{red!35}0.196 & \cellcolor{red!35}0.63 & \cellcolor{red!35}0.08 & \cellcolor{red!35}0.57 \\

CBDBuilding & 2.47 & 0.94 & 0.12 & $\times$ & \cellcolor{orange!35}0.78 & \cellcolor{orange!35}0.74 & \cellcolor{red!35}0.06 & \cellcolor{red!35}0.60 & \cellcolor{red!35}0.49 & \cellcolor{red!35}0.64 & \cellcolor{red!35}0.06 & \cellcolor{red!35}0.60 \\

Company & 1.04 & 0.45 & \cellcolor{orange!35}0.064 & $\times$ & \cellcolor{orange!35}0.29 & \cellcolor{orange!35}0.41 & 0.07 & \cellcolor{orange!35}0.54 & \cellcolor{red!35}0.28 & \cellcolor{red!35}0.40 & \cellcolor{red!35}0.06 & \cellcolor{red!35}0.49 \\

Street  & 1.24 & 0.56 & 0.079 & $\times$ & \cellcolor{orange!35}0.28 & \cellcolor{orange!35}0.49 & \cellcolor{orange!35}0.064 & \cellcolor{orange!35}0.52 & \cellcolor{red!35}0.24 & \cellcolor{red!35}0.46 & \cellcolor{red!35}0.058 & \cellcolor{red!35}0.5 \\
\bottomrule

\end{tabular}
\end{threeparttable}
\vspace{-0.3cm}
\end{table*}


\subsection{Robustness and Sensitivity Analysis}
\begin{figure}[!t]
  \centering
  \includegraphics[width=1.0\linewidth]{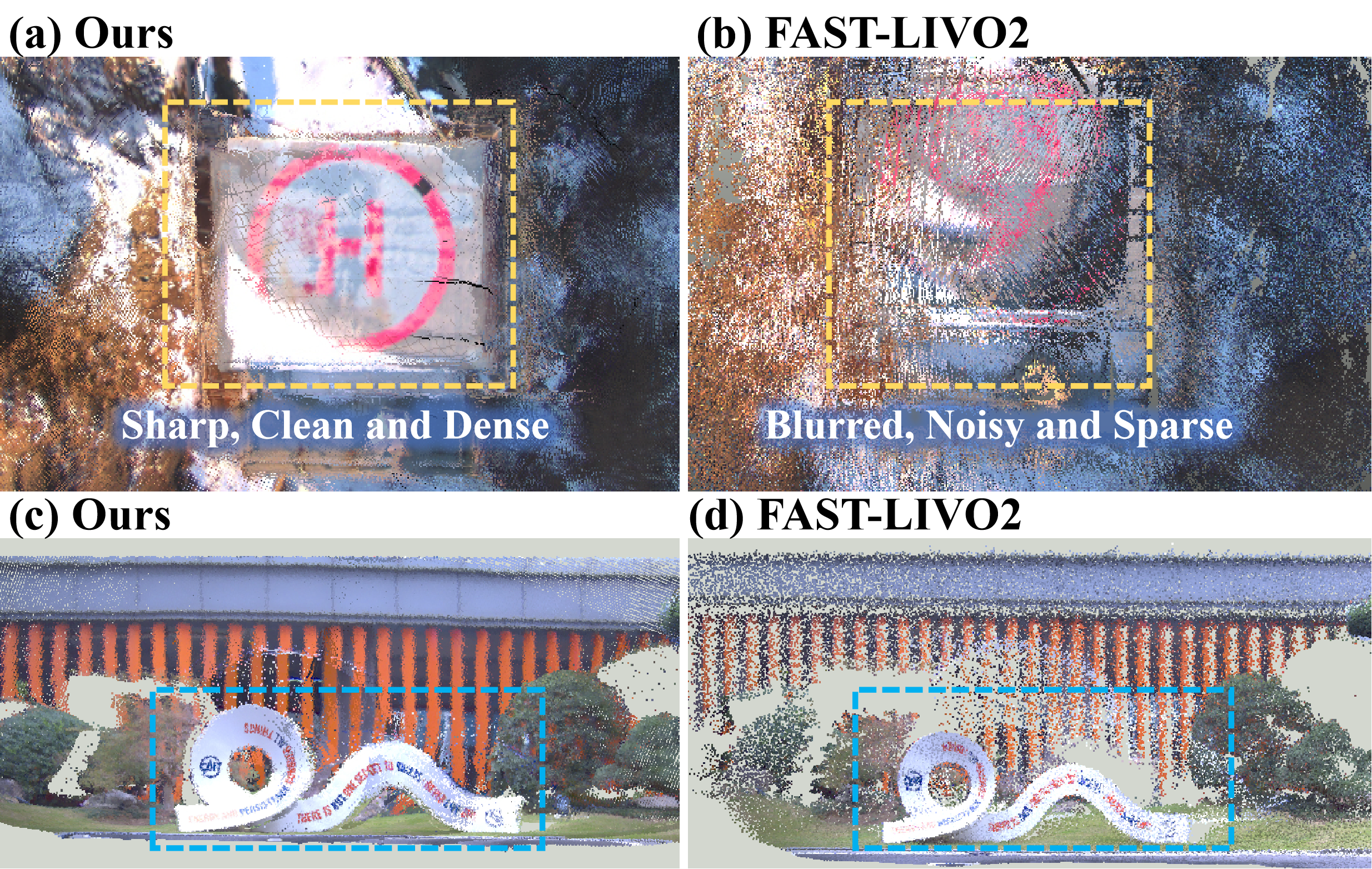}
   \vspace{-20pt}
  \caption{(a) and (b) compare our method with FAST-LIVO2 on MUN\_FRL, while (c) and (d) show mapping results on our dataset TechnologyPark. Benefiting from the correspondence between points and pixels, our reconstructions are significantly denser, sharper, and cleaner.}
  \label{fig:sensitivity}
  \vspace{-0.3cm}
\end{figure}

We evaluate our method and FAST-LIVO2 on the self-collected and MUN\_FRL datasets. The former features roughly calibrated extrinsic parameters but reliable hardware time synchronization, whereas the latter offers accurately calibrated extrinsics but lacks time synchronization. Fig.~\ref{fig:sensitivity} compares the mapping results of the two methods. It can be observed that in the absence of either hardware time synchronization or precise extrinsic calibration, FAST-LIVO2 suffers significant performance degradation, producing blurred point clouds. In contrast, our method remains largely unaffected, producing sharper, cleaner, and denser point clouds. More visualization can be found in the supplementary video.

\section{Conclusion}
This paper presents LiDAR-VGGT, a novel framework that integrates LiDAR with VGGT to achieve large-scale, dense, metric-accurate, and globally consistent reconstruction.
We design a pre-fusion module for robust initialization and a post-fusion module featuring enhanced cross-modal registration, enabling LiDAR inertial odometry to effectively refine VGGT poses and maps in a coarse-to-fine manner.
To our best knowledge, this is the first work to combine LiDAR and VGGT, demonstrating that LiDAR fusion can successfully recover the real-world metric attributes of VGGT reconstructions.
Nonetheless, the current fusion framework does not directly incorporate LiDAR information into the transformer backbone, which limits the tightness of cross-modal coupling. As future work, we plan to embed LiDAR cues directly into the VGGT pipeline, enabling tighter integration and end-to-end cross-modal reconstruction with improved scalability, efficiency, and robustness.


\bibliographystyle{IEEEtran}
\bibliography{ref}
\vfill

\end{document}